\documentclass{article}
\usepackage{ijcai22}

\usepackage{times}
\usepackage{soul}
\usepackage{url}
\usepackage[hidelinks]{hyperref}
\usepackage[utf8]{inputenc}
\usepackage[small]{caption}
\usepackage{graphicx}
\usepackage{amsmath}
\usepackage{amsthm}
\usepackage{booktabs}
\usepackage{algorithm}
\usepackage{algorithmic}
\title{S-SimCSE: Sampled Sub-networks for Contrastive Learning of Sentence Embedding}

\author{
Junlei Zhang$^1$
\and
Zhenzhong Lan$^1$
\affiliations
$^1$Westlake University
\emails
$^1$zhangjunlei@westlake.edu.cn
}

\begin{document}

\maketitle

\begin{abstract}
Contrastive learning has been studied for improving the performance of learning sentence embeddings. The current state-of-the-art method is the SimCSE, which takes dropout as the data augmentation method and feeds a pre-trained transformer encoder the same input sentence twice. The corresponding outputs, two sentence embeddings derived from the same sentence with different dropout masks,  can be used to build a positive pair. A network being applied with a dropout mask can be regarded as a sub-network of itsef, whose expected scale is determined by the dropout rate. In this paper, we push sub-networks with different expected scales learn similar embedding for the same sentence. SimCSE failed to do so because they fixed the dropout rate to a tuned hyperparameter. We achieve this by sampling dropout rate from a distribution eatch forward process. As this method may make optimization harder, we also propose a simple sentence-wise mask strategy to sample more sub-networks. We evaluated the proposed S-SimCSE on several popular semantic text similarity datasets. Experimental results show that S-SimCSE outperforms the state-of-the-art SimCSE more than $1\%$ on BERT$_{base}$

\end{abstract}

\section{Introduction}
How to learn universal sentence embeddings by large-scale pre-trained models\cite{lan2019albert,devlin2018bert}, such as BERT, has been studied extensively in the literature \cite{reimers2019sentence,gao2021simcse}. Recently, researchers have proposed using contrastive learning to learn better sentence embeddings. Generally, contrastive learning use various data augmentation methods to generate different views of the input sentences and force the models to learn more semantically similar embeddings with their augmented counterpart than other sentences. The current state-of-the-art method is SimCSE\cite{gao2021simcse}, which uses the dropout as the data augmentation method. Specifically, SimCSE feeds a pre-trained language model the same sentence twice with two independently sampled dropout masks. Then the embeddings derived from the same input sentence are regarded as the positive embedding pairs. Other sentences are regarded as negative sentences. 

Dropout works by applying a dropout mask to a network according to the dropout rate. which can be regarded as sampling a sub-network of itself. In the SimCSE, they adopted a fixed dropout rate during the training process. This means that the expected scales of the sampled sub-networks are the same. In this paper, we proposed a simple and effective way to advance the SimCSE. Instead of using the fixed dropout rate, we sampled different dropout rates for each of the dropout operations in each of the forward process. Intuitively, for input sentences, we push two sub-networks sampled with different dropout rates to learn similar embeddings. 
Furthermore, to sample more sub-networks for better optimization, we sample dropout rates separately for each of the input sentences in the same batch.

\section{Background}


SimCSE applied contrastive learning \cite{chen2020simple} on the universal sentence learning problem and the dropout is used as the data augmentation function. Specially, given a pair of input sentences $\left\{x_i, x_i^{+}\right\}$, where $x_i$ and $x_i$ are two semantically related sentence pairs and $i$ is the index of the sentence in a mini-batch. let $h_i$ and $h_i^{+}$ denotes the projected embeddings of $x_i$ and $x_i^{+}$ by a encoder function$f_\theta$. For a mini-batch with N pairs, the training loss for is $\left\{x_i, x_i^{+}\right\}$:
\begin{equation}
    \label{simclrloss}
    l_i =-log\frac{e^{sim(h_i, h_i^{+}) / \tau} }{\sum_{j=1}^{N} e^{sim(h_i, h_j) / \tau}}
\end{equation}
, where $\tau$ is a temperature parameter and the $sim(\cdot, \cdot)$ is typically the cosine similarity function as follows:
\begin{equation}
    sim(h_i, h_i^{+})=\frac{h_i^{T}h_i^{+}}{\left \| h_i \right \|\left \| h_i^{+} \right \|}
\end{equation}

The core idea of SimCSE is to use different dropout masks to build positive pairs. In the supervised setting, the positive pair is composed of  $x_i$ and $x_i^{j}$ two different but semantically similar sentences based on their labels. The dropout function is applied to the encoder network when projecting $x$to $h$. In the unsupervised setting, the input sentence $x_i$ is fed into the encoder twice by applying different dropout masks $m_i$ and $m_i^{+}$  separately. Then the positive pair embeddings can be got by: Uniform distribution
\begin{equation}
    h_i=f_{\theta}(x_i, m_i), h_i^{+}=f_{\theta}(x_i,m_i^{+})
\end{equation}
Then $h_i$ and $h_i^{+}$ can be used as positive pair in Equ. \ref{simclrloss}.

\section{Method}
In this section, we first introduce the dropout rate sampling method to sample sub-networks for better contrastive learning. Then we introduce the individual dropout rate sampling strategy for sampling more sub-networks.
\subsection{Dropout rate sampling}
The dropout function randomly masks a network to get a sub-network and the expected scale of the sub-network is determined by the dropout rate. In the original SimCSE, they use the same dropout rate but the different dropout mask to get $h_i$ and $h_i^{+}$. Instead of using fix dropout rate, we sample dropout rates from a pre-defined distribution. Specifically, we firstly sample two dropout rates $r_1$ and $r_2$ from a pre-defined distribution (e.g. uniform distribution). Then, following the SimCSE, we feed the input sentence $x_i$ into the network twice, where dropout rates are $r_1$ and $r_2$ separately.  
\subsection{Sentence-wise dropout mask}
Srivastava et al. \cite{srivastava2014dropout} proposed applied a random mask on a network during the training process. In this paper, we focus on transformer based architecture (e.g. BERT \cite{devlin2018bert}, Roberta\cite{liu2019roberta} ), where dropout is only used before fully connected layer. Specifically, let $z_i^l$ denotes the vector of outputs for the $i th$ sentence in a mini-batch from layer $l$ (fully connected layer). $w^{l}$ and $b^{l}$ are the weights and biases at the layer l. The feed-forward operation of a standard fully connection layer can be described as (for $l\in \left \{ 0,...,L-1 \right \}$ and $i\in \left \{ 0,...,N-1 \right \}$):
\begin{align}
    \nonumber p &\sim \pi (\tau) \\
    \nonumber m^{l+1} &\sim Bernoulli(p),\\
    \nonumber x^{l} &= m^{l+1} * x^{l},\\
    z^{l+1} &= w^{l+1}x^{l} + b^{l+1}
\end{align}
for each of the sentence in a mini-batch, we sample a new mask with dropout rate sampled from a distribution $\tau$. Thus, we can get sub-networks with different dropout rate in a single forward process.

\section{Experiments}
\subsection{Evaluation Setup}
Following SimCSE \cite{gao2021simcse}, we conduct our experiments on 7 standard semantic textual similarities (STS) tasks. All the STS experiments are fully unsupervised and no STS training sets are used. Even for the supervised setting, we simply take extra labeled datasets instead of STS training sets for training. Specifically, for the unsupervised setting, we use 1-million sentences randomly drawn from English Wikipedia and directly evaluate our model on the test set of 7 STS tasks.

\subsection{Semantic Textual Similarity Tasks}
Semantic textual similarity measures the semantic similarity of two sentences. STS 2012-2016 \cite{agirre2013sem,agirre2014semeval,agirre2015semeval,agirre2016semeval} and STS-B \cite{cer2017semeval} are widely used evaluation benchmark for semantic textual similarity tasks. Following SimCSE, we measure the semantic similarity of two sentences with the cosine similarity. After deriving the semantic similarities of all pairs in the test set, we follow SimCSE to use Spearman correlation to measure the correlation between the ranks of predicted similarities and the ground truth. For a set of size $n$, the $n$ scores of $X_i, Y_i$ are converted to its corresponding ranks $rg_{X_i), rg_{Y_i}}$, then the Spearman correlation is defined as follows:
\begin{equation}
    r_s=\frac{cov(rg_X, rg_Y)}{\sigma _{rg_X}\sigma_{rg_Y}}
\end{equation}
where $conv(rg_X, rg_Y)$ is the covariance of the rank variables,  $\sigma _{rg_X} and \sigma_{rg_Y}$ are the standard deviations of the rank variables. Spearman correlation has a value between -1 an 1, which will be high when the ranks of predicted similarities and the groundtruth are similar.

\subsection{Main Results}
\begin{table}[h!]
\centering
\caption{The average and max of sentence embedding performances on 7 semantic textual similarities (STS) test sets, in terms of Spearman's correlation, with different models. The Avg. results are calculated from the top 3 results of seven different seeds. $w/$ means the sentence-wise dropout mask is applied on top of the dropout rate sampling strategy.}

\begin{tabular}{@{}lcc@{}}

\toprule
\textbf{Model}                    & \textbf{Avg.} & \textbf{Max} \\ \midrule
SimCSE-BERT$_{base}$           & $75.99\pm 0.38$   & 76.18        \\
S-SimCSE-BERT$_{base}$ w/o  sd   & $76.82\pm 0.20$     & 77.08        \\
S-SimCSE-BERT$_{base}$ w/  sd      & $76.92\pm  0.32  $  & \textbf{77.35}        \\
SimCSE-BERT$_{large}$            & $78.15 \pm  0.22 $  & 78.42        \\
S-SimCSE-BERT$_{large}$ w/o  sd    & $78.67\pm  0.33  $  & 78.99        \\
S-SimCSE-BERT$_{large}$ w/  sd     & $79.27 \pm 0.23$    & \textbf{79.42}        \\
SimCSE-Roberta$_{base}$         & $ 76.87 \pm 0.17 $   & 77.03        \\
S-SimCSE-Roberta$_{base}$ w/o  sd & $77.03  \pm  0.18 $ & 77.24        \\
S-SimCSE-Roberta$_{base}$ w/  sd   & $77.08\pm 0.26 $  & \textbf{77.33}        \\ \bottomrule
\end{tabular}
\end{table}

\section*{Acknowledgments}

\bibliographystyle{named}
\bibliography{ijcai22}

\end{document}